\documentclass{article}
\usepackage{ijcai19}

\usepackage{times}
\usepackage{soul}
\usepackage{url}
\usepackage[hidelinks]{hyperref}
\usepackage[utf8]{inputenc}
\usepackage[small]{caption}
\usepackage{graphicx}
\usepackage{amsmath}
\usepackage{booktabs}
\usepackage{algorithm}
\usepackage{algorithmic}
\usepackage{booktabs}
\usepackage{tabulary}
\usepackage{enumitem}
\usepackage{setspace}
\usepackage{amsfonts}
\usepackage{subfigure}
\usepackage[table,x11names]{xcolor}

\newcommand{\tbf}{\textbf}

\title{Generative Visual Dialogue System via Weighted Likelihood Estimation}

\author{
Heming Zhang$^1$\footnote{Contact Author}\and
Shalini Ghosh$^2$\and
Larry Heck$^2$\and
Stephen Walsh$^2$\and \\
Junting Zhang$^1$\and
Jie Zhang$^3$ \And
C.-C. Jay Kuo$^1$
\affiliations
$^1$University of Southern California\\
$^2$Samsung Research America\\
$^3$Arizona State University
\emails
\{hemingzh, juntingz, cckuo\}@usc.edu,
\{shalini.glosh, larry.h, s1.walsh\}@samsung.com,
jiezhang.joena@asu.edu
}

\begin{document}

\maketitle

\begin{abstract}

The key challenge of generative Visual Dialogue (VD) systems is to respond to human queries with informative answers in natural and contiguous conversation flow. Traditional Maximum Likelihood
Estimation-based methods only learn from positive responses but ignore the negative responses, and consequently tend to yield safe or generic responses.
To address this issue, we propose a novel training scheme in conjunction with weighted likelihood estimation method.
Furthermore, an adaptive multi-modal reasoning module is designed, to accommodate various dialogue scenarios automatically and select relevant information accordingly. 
The experimental results on the VisDial benchmark demonstrate the superiority of our proposed algorithm over other state-of-the-art approaches, with an improvement of 5.81\% on recall@10.
\end{abstract}

\section{Introduction}
Artificial Intelligence (AI) has witnessed rapid resurgence in recent years, due to many innovations in deep learning. Exciting results have been obtained in computer vision ({\em e.g.}, image classification \cite{VGG,ResNet}, detection \cite{faster_rcnn,RetinaNet,face}, etc.) as well as natural language processing (NLP) ({\em e.g.}, \cite{network_based,adversarial_learning,zhangjie}, etc.). Good progress has also been made by researchers in vision-grounded NLP tasks such as image captioning \cite{image_captioning,visual_genome} and visual question answering~\cite{vqa,ask_your_neurons}. Proposed recently, the Visual Dialogue (VD)~\cite{VD} task leads to a higher level of interaction between vision and language. In the VD task, a machine conducts natural language dialogues with humans by answering questions grounded in an image. It requires not only reasoning on vision and language, but also generating consistent and natural dialogues. 

%%%%%%%%%%%%%%%%%%%%%%%%%%%%%%%%%%%%%%%%%%
\begin{figure}
\centering
\addtolength{\subfigcapskip}{-0.2in}
\addtolength{\subfigbottomskip}{-0.2in}
\subfigure[]{
\label{fig:example}
\includegraphics[trim=170 130 910 280,clip,scale=0.3]{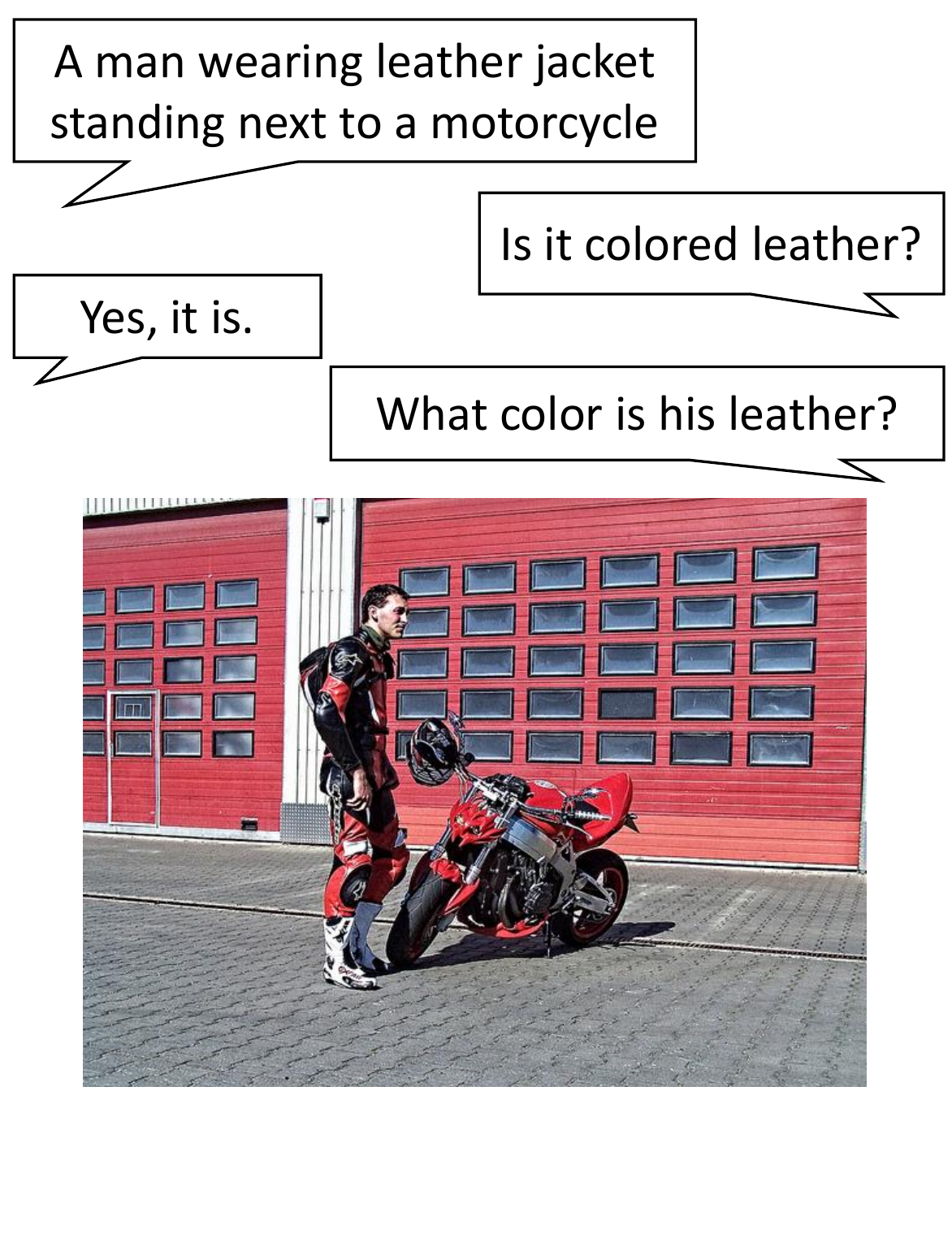}}
\subfigure[]{
\label{fig:comparison}
\includegraphics[trim=520 130 490 280,clip,scale=0.3]{figures/comparison.pdf}}
\vspace{-0.05in}
\caption{(a) An example from the VisDial dataset, and (b) comparison
between MLE, GAN and WLE, where positive responses are highlighted in
blue.  The MLE-based generator learn from data in positive answers only.
The GAN-based generator learn from data in negative answers through
discriminators indirectly.  Our WLE-based generator learn from data in
both positive and negative answers.}
\end{figure}
%%%%%%%%%%%%%%%%%%%%%%%%%%%%%%%%%%%%%%%%%%

Existing VD systems can be summarized into two tracks
\cite{VD}:
generative models and discriminative models. The system adopting the generative model can generate responses while that using the disriminative model only chooses responses from a candidate set.
Although discriminative models achieved better recall performance on the benchmark dataset~\cite{VD}, they are not as applicable as generative models in real world scenarios since candidate responses may not be available. In this work, we focus on the design of generative VD systems
for broader usage. 

One main weakness of existing generative models trained by the maximum likelihood estimation (MLE) method is that they tend to provide frequent and generic responses like `Don't know'. This happens because the MLE training paradigm latches on to frequent generic responses~\cite{best_of_both_worlds}. They may match well with some but poorly for others. There are many possible paths a dialogue may take in the future --- penalizing generic poor responses can eliminate candidate dialogue paths and avoids abuse of frequent responses. This helps bridge the large performance gap between generative/discriminative VD systems. 

To reach this goal, we propose a novel weighted likelihood estimation~(WLE) based training scheme.
Specifically, instead of assigning equal weights to each training sample as done in the MLE, we assign a different weight to each training sample. The weight of a training sample is determined by its positive response as well as the negative ones. By incorporating supervision from both positive and negative responses, we enhance answer diversity in the resulting generative model. The proposed training scheme is effective in boosting the VD performance and easy to implement. 

Another challenge for VD systems is effective reasoning based on multi-modal inputs. Previous work pre-defined a set of reasoning paths based on multi-modal inputs. The path is specified by a certain sequential processing order, {\em e.g.}, human queries followed by the dialogue history and then followed by image analysis \cite{best_of_both_worlds}. Such a pre-defined order is not capable of handling different dialogue scenarios, e.g., answering a follow-up question of `Is there anything else on the table?'. 
We believe that a good reasoning strategy should determine the processing order by itself. 
Here, we propose a new reasoning module, where an adaptive reasoning path accommodates different dialogue scenarios automatically. 

There are three major contributions of this work. 
First, an effective training scheme for the generative VD system is proposed, which directly exploits both positive and negative responses using an unprecedented likelihood estimation method. Second, we design an adaptive reasoning scheme with unconstrained attention on multi-modal inputs to accommodate different dialogue scenarios automatically. Third, our results demonstrate the state-of-the-art performance on the VisDial dataset~\cite{VD}.
Specifically, our model outperforms the
best previous generative-model-based method \cite{are_you_talking_to_me}
by 3.06\%, 5.81\% and 5.28 with respect to the recall@5, the recall@10
and the mean rank performance metrics, respectively. 

\section{Related Work}\label{sec:related}

\paragraph{Visual dialogue.} Different visual dialogue tasks have been
examined recently. The VisDial dataset~\cite{VD} is collected from
free-form human dialogues with a goal to answer questions related to a
given image. The GuessWhat task~\cite{guess_what} is a
guessing game with goal-driven dialogues so as to identify a certain
object in a given image by asking yes/no questions. In this work, we focus on the VisDial task. 
Most previous research on the VisDial task follows the encoder-decoder
framework in~\cite{encoder_decoder}. Exploration on encoder models
includes late fusion~\cite{VD}, hierarchical recurrent
network~\cite{VD}, memory network~\cite{VD}, history-conditioned image
attentive encoder~(HCIAE)~\cite{best_of_both_worlds}, and sequential
co-attention~(CoAtt)~\cite{are_you_talking_to_me}.  Decoder models can
be classified into two types: (a) Discriminative decoders rank candidate
responses using cross-entropy loss~\cite{VD} or n-pair
loss~\cite{best_of_both_worlds}; (b) Generative decoders yield responses
using MLE~\cite{VD}, which can be further combined with adversarial
training~\cite{best_of_both_worlds,are_you_talking_to_me}. The latter
involves a discriminator trained on both positive and negative
responses, and its discriminative power is then transferred to the
generator via auxiliary adversarial training.  

\paragraph{Weighted likelihood estimation.} Being distinct from previous generative work that uses either MLE or adversarial training, we use WLE and develop a new training scheme for VD systems in this work. WLE has been utilized for different purposes. For example, it was introduced in~\cite{WLE_bias} to remove the first-order bias in MLE. Smaller weights are assigned to outliers for training to reduce the effect of outliers~\cite{WLE_robust}.  The binary indicator function and the similarity scores are compared for weighting the likelihood in visual question answering (VQA) in~\cite{WLE_VQA1}. We design a novel weighted likelihood remotely related to these concepts, to utilize both positive and negative responses.

\paragraph{Hard example mining.} Hard example mining methods are frequently seen in object detection algorithms, where the amount of background samples is much more than the object samples. In~\cite{offline_HEM}, the proposed face detector is trained until convergence on sub-datasets and applied to more data to mine the hard examples alternatively. Online hard example mining is favored by later work~\cite{OHEM,RetinaNet}, where the softmax-based cross entropy loss is used to determine the difficulty of samples. We adopt the concept of sample difficulty and propose a novel way to find hard examples without the preliminary of softmax-based cross entropy.

\paragraph{Multi-modal reasoning.} Multi-modal reasoning involves
extracting and combining useful information from multi-modal inputs.  It
is widely used in the intersection of vision and language, such as image
captioning~\cite{caption_attention_1} and VQA~\cite{vqa_attention_1}.
For the VD task, reasoning can be applied to images~(I), questions~(Q)
and history dialogues~(H).  In~\cite{best_of_both_worlds}, the reasoning
path adopts the order ``Q~$\rightarrow$~H~$\rightarrow$~I". This order
is further refined to ``Q~$\rightarrow$~I~$\rightarrow$~H~
$\rightarrow$~Q" in~\cite{are_you_talking_to_me}. 
In the recent arxiv paper~\cite{ReDAN}, the reasoning sequence of ``Q~$\rightarrow$~I~$\rightarrow$~H~" is recurrently occurring to solve complicated problems.
Unlike previous work
that defines the reasoning path order a priori, we propose an adaptive
reasoning scheme with no pre-defined reasoning order. 

\section{Proposed Generative VD System}

In this section, we describe our approach to construct and train the
proposed generative visual dialogue system.  Following the problem
formulation in \cite{VD}, the input consists of an image $I$, a
`ground-truth' dialogue history $H_{t-1}=(\underbrace{C}_{h_0},
\underbrace{(Q_1,A_1)}_{h_1}, \cdots,
\underbrace{(Q_{t-1},A_{t-1})}_{h_{t-1}})$ with image caption $C$ and a
follow-up question $Q_t$ at round $t$.  $N$ candidate responses
$\mathcal{A}_t = \{A_t^1, A_t^2, \cdots, A_t^{N}\}$ are provided for
both training and testing. Figure~\ref{fig:example} shows an example
from VisDial~\cite{VD}. 

We adopt the encoder-decoder framework~\cite{encoder_decoder}.  Our
proposed encoder, which involves an adaptive multi-modal reasoning
module without pre-defined order, will be described in details in
Sec.~\ref{sec:attention}.  The generative decoder receives the embedding
of the input triplet $\{I, H_{t-1}, Q_t\}$ from the encoder and outputs
a response sequence $\hat{A}_t$. Our VD system is trained using a novel
training scheme with weighted likelihood estimation, which will be
described in Sec.~\ref{sec:wle} with details. 

\subsection{Adaptive Multi-modal Reasoning~(AMR)}\label{sec:attention}
To conduct reasoning on multi-modal inputs, we first extract image feature $F_{I} \in \mathbb{R}^{N \times H \times W}$ by a convolutional neural network, where $N$ is the length of the feature, and $H$ and $W$ are the height and width of the output feature map. The question feature $F_Q \in \mathbb{R}^{N \times l_Q}$ and history feature $F_H \in \mathbb{R}^{N \times l_H}$ are obtained by recurrent neural networks, where $l_Q$ and $l_H$ are the length of the question and the history, respectively. 

% attention
Our reasoning path consists of two main steps, namely the comprehension step and the exploration step, in a recurrent manner. In the comprehension step, useful information from each input modality is extracted. It is apparent that not all the input information is equally important in the conversation. Attention mechanism is thus useful to extract relevant information.
In the exploration step, the relevant information is processed and the following attention direction is determined accordingly. Along the reasoning path, these two steps are performed alternatively.

In~\cite{best_of_both_worlds,are_you_talking_to_me}, 
%the comprehension and exploration steps are alternatively performed following a pre-defined reasoning sequence though each input modality.
the comprehension and exploration steps are merged together. The reasoning scheme focuses on one single input modality at each time and follows a pre-defined reasoning sequence through each input modality.
However, this pre-defined order cannot accommodate various dialogue scenarios in real world. For example, a question of ``How many people are there in the image?" should yield a short reasoning sequence like 
\begin{equation*}
\underset{\text{the word `people'}}{\texttt{\small question}}  \rightarrow \underset{\text{regions of people}}{\texttt{\small image}},
\end{equation*}
whereas a question of ``Is there anything else on the table?" should result in a long reasoning sequence such as
\begin{equation*}
\underset{\text{the word `table'}}{\texttt{\small question}}  \rightarrow
\underset{\text{regions of table}}{\texttt{\small image}}  \rightarrow
\underset{\text{the word `else'}}{\texttt{\small question}}  \rightarrow
\underset{\text{context for `else'}}{\texttt{\small history}}.
\end{equation*}

\begin{figure}
\centering
\includegraphics[trim=360 210 430 100,clip,width=0.75\columnwidth]{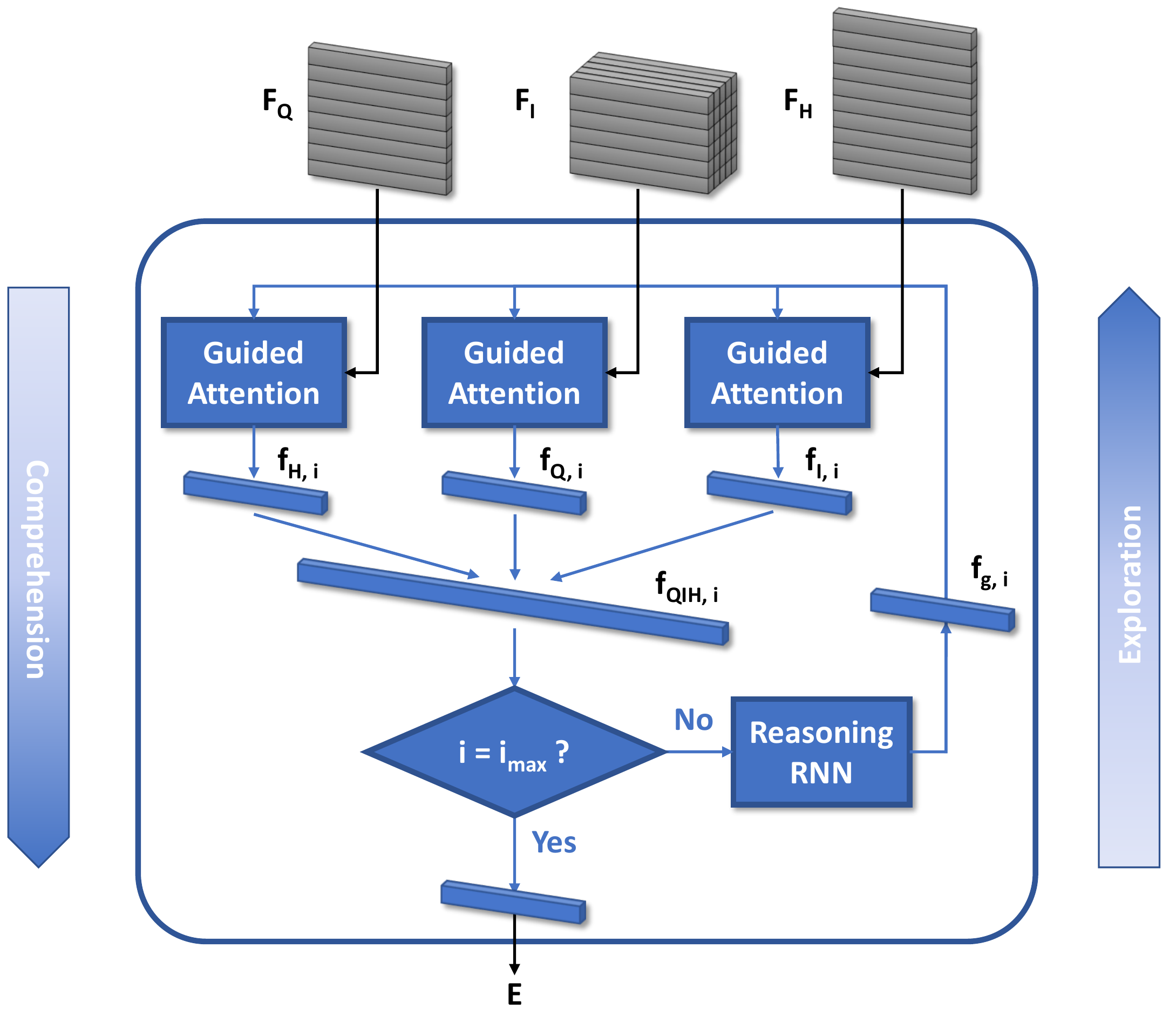}
\caption{The adaptive multi-modal reasoning.}
\label{fig:recurrent_attention}
\end{figure}

To overcome the drawback of pre-defined reasoning sequence, we propose an adaptive multi-modal reasoning module as illustrated in Figure \ref{fig:recurrent_attention}. 

%%%%%%%%%%%%%%%%%%%%%%%%%%%%%%%%%%%%%%%%%%
%%%%%%%%%%%%%%%%%%%%%%%%%%%%%%%%%%%%%%%%%%

Let $\lambda$ denote any multi-modal feature type (image, question or history), and $F_\lambda \in \mathbb{R}^{N \times M}$ denote the features to be attended, where $M$ is the number of features. The guided attention operation that paying attention according to the given guide
is denoted as $\mathbf{f}_\lambda = \text{GuidedAtt}(F_\lambda, \mathbf{f}_g)$, where $\mathbf{f}_g \in \mathbb{R}^{N \times 1}$ is the attention guiding feature. The guided attention can be expressed as:
\begin{align}
E_{\lambda} &= \tanh(W_\lambda F_\lambda + W_g \mathbf{f}_g \mathbf{1}^T)\\ % L x N
\mathbf{a}_\lambda &= \text{softmax}(E_{\lambda}^T \mathbf{w}_{att})\\ % N x 1
\mathbf{f}_\lambda &= F_\lambda \mathbf{a}_\lambda,
\end{align}
where $W_\lambda$, $W_g$ and $\mathbf{w}_{att}$ are learnable weights, $\mathbf{1}$ is a vector with all elements set to 1.

In time step $i$, the image features $F_I$, the question features $F_Q$ and the history features $F_H$ are attended separately by their own guided attention blocks. During the comprehension step, the outputs of the guided attention blocks $\mathbf{f}_{I,i}$, $\mathbf{f}_{Q,i}$ and $\mathbf{f}_{H,i}$, {\em i.e.} the extracted information from each modality, are merged into $\mathbf{f}_{QIH,i}$. During the exploration step, the merged vector is processed in the reasoning RNN block, which generates the new attention guiding feature $\mathbf{f}_{g,i}$ to guide the attention in time step $i+1$. 
The final embedding feature $\mathbf{E}$ is
\begin{equation}
\mathbf{E} = \tanh(W \mathbf{f}_{QIH,i_{max}}),
\end{equation}
where $W$ is learnable weights, $i_{max}$ is the maximum number of recurrent steps.

Through this mechanism, the reasoning RNN block maintains a global view of the multi-modal features and reasons what information should be extracted in the next time step. The information extraction order and subject are therefore determined adaptively along the reasoning path.

\subsection{WLE-based Training Scheme}
\label{sec:wle}
% MLE
As the discriminative VD models are trained to differentiate positive and negative responses, they perform better on the standard discriminative benchmark. In contrast, the generative visual dialogue models are trained to only maximize the likelihood of positive responses. The MLE loss function is expressed as:
\begin{equation}
L_{MLE} = \textstyle \sum_m - \log(p^{pos}_m),
\end{equation}
where $p^{pos}_m$ denotes the estimated likelihood of the positive response of sample $m$. There is only one positive response per sample provided for training in the VisDial task. However, there are many possible paths a dialogue may take in the future, the MLE approach therefore favors the frequent and generic responses when the training data is limited \cite{best_of_both_worlds}. In the VisDial task, negative responses are selected from positive responses to other questions, including frequent and generic responses. Incorporating the negative responses to maximize the learning from all available information is thus essential to improve the generative models.

We propose a WLE based training scheme to utilize the negative responses and remedy the bias of MLE. Rather than treating each sample with equal importance, we assign a weight $\alpha_m$ to each estimated log-likelihood as:
\begin{equation}
L_{WLE} = \textstyle \sum_m - \alpha_m \log(p^{pos}_m).
\end{equation}
We can interpret the weighted likelihood as a hard sample mining process. We are inspired by OHEM \cite{OHEM} and focal loss \cite{RetinaNet} designed for object detection, where hard samples are mined using their loss values and receive extra attention. Rather than using the preliminary softmax cross entropy loss for discriminative learning, we propose to use likelihood estimation to mine the hard samples.
If the current model cannot predict the likelihood for a sample well, it indicates that this sample is hard for the model. Then we should increase the weight for this hard sample and vice versa.

Given both positive and negative responses for training, we propose to assign weights as:
\begin{align}
\beta_{m,n} &=  1-\frac{\log(p^{neg}_{m,n})}{\log(p^{pos}_m)}, \\
\tilde{\beta}_m &= \exp\left(\tau  \max_n \left( \beta_{m,n}\right) \right),\\
%\frac{1}{|\mathcal{A}_m^{neg}|} \sum_n \exp\left(\tau \beta_{m,n}\right), \\
\alpha_m &= \max\left(\tilde{\beta}_m, \; \gamma\right),
\end{align}
%where $\mathcal{A}_m^{neg}$ denotes the set of negative responses of sample $m$, 
where $p^{neg}_{m,n}$ denotes the $n$-th negative response of sample $m$, $\tau$ and $\gamma$ are hyper-parameters to shape the weights.

We can also view the proposed loss function as a ranking loss. We assign a weight to a sample by comparing the estimated likelihood of its positive and negative responses. $\beta_{m,n}$ measures the relative distance of likelihood between the positive response and the $n$-th negative response of sample $m$. If the likelihood of a positive response is low comparing to the negative responses, we should penalize more by increasing the weight for this sample. If the estimated likelihood of a positive sample is already very high, we should lower its weight to reduce the penalization.

%%%%%%%%%%%%%%%%%%%%%%%%%%%%%%%%%%%
\begin{table}
\centering
\singlespacing
\tabcolsep=0.11cm
\scalebox{0.9}{
\begin{tabulary}{\columnwidth}{LCCCCC}
\toprule
Model & MRR & R@1 & R@5 & R@10 & Mean\\
\midrule
\footnotesize LF~\scriptsize{\cite{VD}} & 0.5199	 & 41.83 & 61.78 & 67.59 & 17.07\\
\footnotesize HREA~\scriptsize{\cite{VD}} & 0.5242 & 42.28 & 62.33 & 68.17 & 16.79\\
\footnotesize MN~\scriptsize{\cite{VD}} & 0.5259 & 42.29 & 62.85 & 68.88 & 17.06\\
\footnotesize HCIAE~\scriptsize{\cite{best_of_both_worlds}} & 0.5467 & 44.35 & 65.28 & 71.55 & 14.23\\
\footnotesize FlipDial~\scriptsize{\cite{flipdial}} & 0.4549 & 34.08 & 56.18 & 61.11 & 20.38\\
\footnotesize CoAtt~\scriptsize{\cite{are_you_talking_to_me}} & 0.5578 & \textbf{46.10} & 65.69 & 71.74 & 14.43\\
%\midrule
\footnotesize Coref~\scriptsize{\cite{visual_coreference_resolution}} & 0.5350 & 43.66 & 63.54 & 69.93 & 15.69\\
\midrule
\small Ours & \textbf{0.5614} & 44.49 & \textbf{68.75} & \textbf{77.55} & \textbf{9.15}\\
\bottomrule
\end{tabulary}}
\vspace{-0.05in}
\caption{Performance of generative models on VisDial 0.9. 
`Mean' denotes mean rank, for which lower is better. 
All the models use VGG as backbone except for Coref which uses ResNet.}
\label{tb:results}
\end{table}
%%%%%%%%%%%%%%%%%%%%%%%%%%%%%%%%%%%

\section{Experiments}
\subsection{Dataset}
\label{sec:experiment_set_up}
We evaluate our proposed model on the VisDial dataset \cite{VD}. 
In VisDial v0.9, on which most previous work has benchmarked, there are in total 83k and 40k dialogues on COCO-train and COCO-val images, respectively. We follow the methodology in \cite{best_of_both_worlds} and split the data into 82k for \texttt{train}, 1k for \texttt{val} and 40k for \texttt{test}. In the new version VisDial v1.0, which was used for the Visual Dialog Challenge 2018, \texttt{train} consists of the previous 123k images and corresponding dialogues. 2k and 8k images with dialogues are collected for \texttt{val} and \texttt{test}, respectively. 

Each question is supplemented with 100 candidate responses, among which only one is the human response for this question. Following the evaluation protocol in \cite{VD}, we rank the 100 candidate responses by their estimated likelihood and evaluate the models using standard retrieval metrics: (1) mean rank of the human response, (2) recall rate of the human response in top-k ranked responses for $k=1,5,10$, (3) mean reciprocal rank (MRR) of the human response, (4) normalized discounted cumulative gain (NDCG) of all correct responses (only available for v1.0).

\subsection{Implementation Details}
\label{sec:training_details}
We follow the procedures in \cite{best_of_both_worlds} to pre-process the data. The captions, questions and answers are truncated at 24, 16 and 8 words for VisDial v0.9, and 40, 20 and 20 words for VisDial v1.0. Vocabularies are built afterwards from the words that occur at least five times in \texttt{train}. We use 512D word embeddings, which are trained from scratch and shared by question, dialogue history and decoder LSTMs.

%%%%%%%%%%%%%%%%%%%%%%%%%%%%%%%%%%%%%
\begin{figure}
\centering
\includegraphics[trim=100 498 555 71,clip,width=0.9\columnwidth]{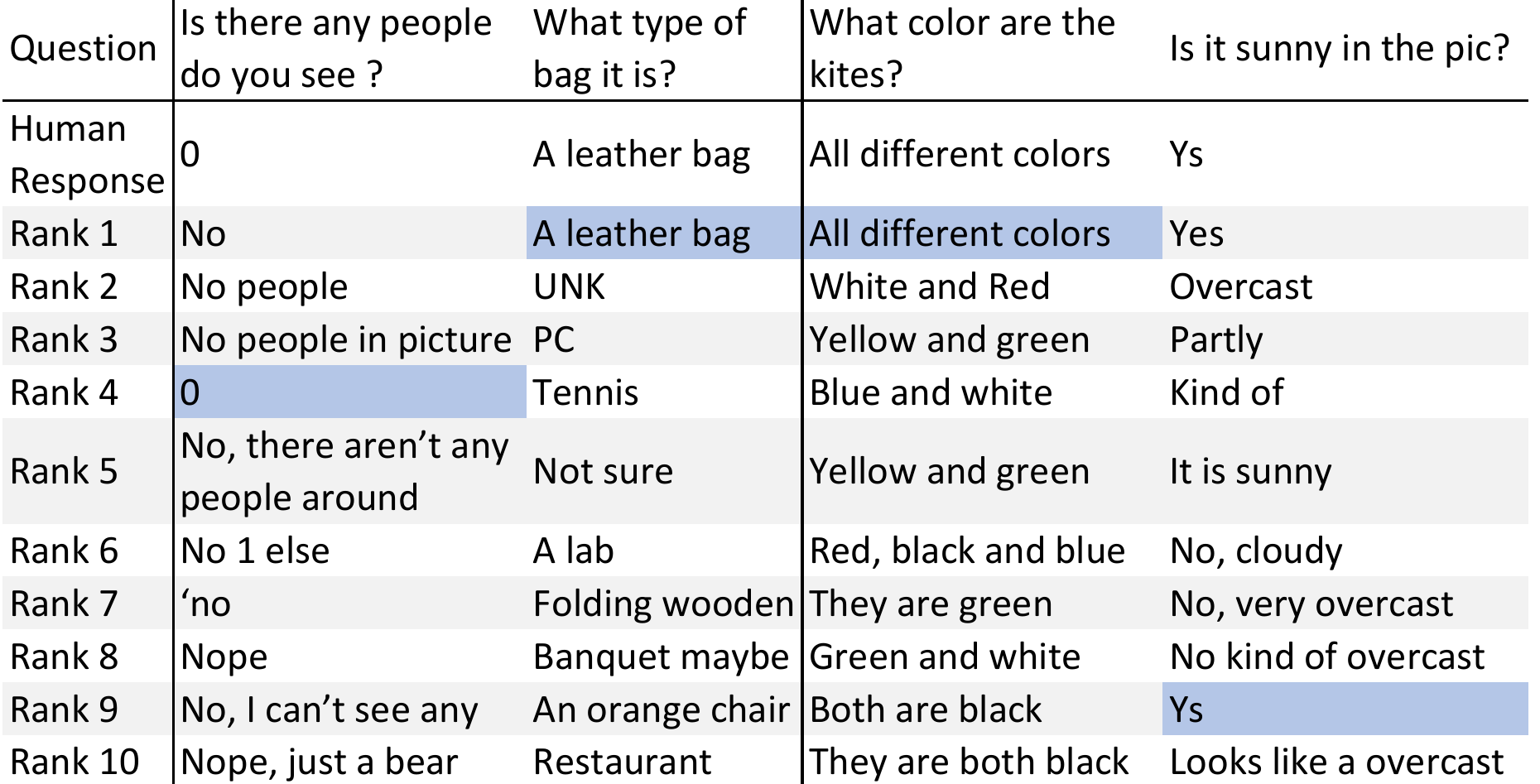}
\caption{Examples of top-10 responses ranked by our model. When there are multiple correct responses to the question, our model may choose other candidates that are semantically similar to the human response. The human responses are highlighted in blue.}
\label{fig:rank_example}
\end{figure}
%%%%%%%%%%%%%%%%%%%%%%%%%%%%%%%%%%%%%

%%%%%%%%%%%%%%%%%%%%%%%%%%%%%%%%%%%
\begin{figure}
\centering
\includegraphics[trim=60 630 210 80, clip,width=0.8\columnwidth]{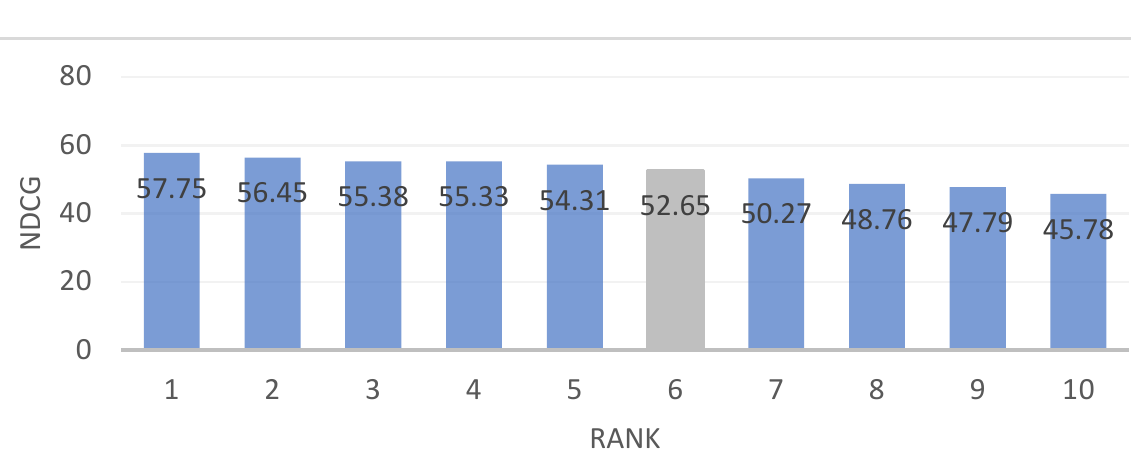}
\caption{Results of the top-10 teams in the first visual dialog challenge. As the only team in top-10 uses generative visual dialogue system, we are ranked as the 6th place (highlighted with gray color). Our NDCG score is comparable with other discriminative systems.}
\label{fig:challenge}
\end{figure}
%%%%%%%%%%%%%%%%%%%%%%%%%%%%%%%%%%%

For a fair comparison with previous work, we adopt the simple LSTM decoder with a softmax output which models the likelihood of the next word given the embedding feature and previous generated sequence. We also set all LSTMs to have single layer with 512D hidden state for consistency with other works. We extract image features from pre-trained CNN models~(VGG~\cite{VGG} for VisDial v0.9, ResNet~\cite{ResNet} or bottom-up features~\cite{bottom_up} for VisDial v1.0), 
and train the rest of our model from scratch. We use the Adam optimizer with the base learning rate of $4 \times 10^{-4}$. 

%%%%%%%%%%%%%%%%%%%%%%%%%%%%%%%%%%%
\begin{table}
\centering
\singlespacing
\tabcolsep=0.11cm
\scalebox{0.9}{
\begin{tabulary}{\columnwidth}{LCCCCC}
\toprule
Model & MRR & R@1 & R@5 & R@10 & Mean\\
\midrule
\footnotesize MN~\scriptsize{\cite{VD}} 
&0.4799      &38.18      &57.54      &64.32      &18.60\\
\footnotesize HCIAE~\scriptsize{\cite{best_of_both_worlds}} 
&0.4910      &39.35      &58.49      &64.70      &18.46\\
\footnotesize CoAtt~\scriptsize{\cite{are_you_talking_to_me}} 
&0.4925      &39.66      &58.83      &65.38      &18.15\\
\footnotesize ReDAN~\scriptsize{\cite{ReDAN}}
&0.4969      &\tbf{40.19}&59.35      &66.06      &17.92\\
\midrule
\small Ours 
&\tbf{0.5015}&38.26      &\tbf{62.54}&\tbf{72.79}&\tbf{10.71}\\
\bottomrule
\end{tabulary}}
\vspace{-0.05in}
\caption{Performance of generative models on VisDial v1.0 val. 
%`Mean' denotes mean rank, for which lower is better. 
Results of previous work are reported by ReDAN.}
\label{tb:v1_0val}
\end{table}
%%%%%%%%%%%%%%%%%%%%%%%%%%%%%%%%%%%

%%%%%%%%%%%%%%%%%%%%%%%%%%%%%%%%%%%%%
\begin{table}
\centering
\tabcolsep=0.11cm
\singlespace
\scalebox{0.9}{
\begin{tabulary}{2\columnwidth}{LCCCCCC}
\toprule
Model & MRR & R@1 & R@5 & R@10 & Mean\\
\midrule
\small HCIAE-MLE
&0.5386 			&44.06 		&63.55 		&69.24 		& 16.01		\\
\small HCIAE-GAN
&0.5467			&44.35		&65.28		&71.55 		& 14.23\\
%\midrule
\small HCIAE-WLE	
&0.5494 			&43.43		&66.88		&75.59		&9.93\\
\small AMR-MLE		
&0.5403 			&44.17		&63.86		&69.67		&15.49\\
\small AMR-WLE		
&\tbf{0.5614}	&\tbf{44.49}	&\tbf{68.75}	&\tbf{77.55}	&\tbf{9.15}\\
\midrule
\midrule
Model & $\Delta$MRR & $\Delta$R@1 & $\Delta$R@5 & $\Delta$R@10 & $\Delta$Mean\\
\midrule
\small HCIAE-MLE
& ---			&---			&---			&---			&--- \\
\small HCIAE-GAN
&+0.0081 		&+0.29 		&+1.73		&+2.31		&-1.78\\
%\midrule
\small HCIAE-WLE
&+0.0108 		&-0.92 		&+3.33 		&+6.35 		&-6.08\\
\small AMR-MLE
& --- 			&---			&---			&--- 		&---\\
\small AMR-WLE
&\tbf{+0.0211}	&\tbf{+0.32}	&\tbf{+4.89}	&\tbf{+7.88}	&\tbf{-6.34}\\
\bottomrule
\end{tabulary}}
\vspace{-0.05in}
\caption{Ablation study on VisDial 0.9. Top: absolute values. Bottom: improvement from MLE models.
%The absolute values are listed on the top and the improvement from MLE models are at the bottom. 
}\label{tb:ablation_results}
\end{table}
%%%%%%%%%%%%%%%%%%%%%%%%%%%%%%%%%%%%%

\subsection{Experiments Results and Analysis}
%\paragraph{Baselines}
\subsubsection{Baselines}
We compare our proposed model to several baselines and the state-of-the-art generative models. In \cite{VD}, three types of encoders are introduced. Late Fusion \textbf{(LF)} extracts features from each input separately and fuses them in the later stage. Hierarchical Recurrent Encoder \textbf{(HRE)} uses hierarchical recurrent encoder for dialogue history and \textbf{HREA} adds attention to the dialogue history on top of the hierarchical recurrent encoder. Memory Network \textbf{(MN)} uses memery bank to store the dialogue history and find corresponding memory to answer the question. 
History-Conditioned Image Attentive Encoder \textbf{(HCIAE)} is proposed in~\cite{best_of_both_worlds} to attend on image and dialogue history and trained with generative adversarial training (GAN). Another concurrent work with GAN~\cite{are_you_talking_to_me} proposes a co-attention model (\textbf{CoAtt}) that attends to question, image and dialogue history. 
\textbf{FlipDial}~\cite{flipdial} uses VAE for sequence generation. We also compare to a neural module network approach \textbf{Coref}~\cite{visual_coreference_resolution} in which only the performance with ResNet~\cite{ResNet} backbone is reported.
\textbf{ReDAN}~\cite{ReDAN} is recently proposed method which involves a multi-step reasoning path with pre-defined order.

%\paragraph{Results on VisDial v0.9}
\subsubsection{Results on VisDial v0.9}
Table \ref{tb:results} compares ours results to other reported generative baselines. Our model performs the best on most of the evaluation metrics. Comparing to HCIAE \cite{best_of_both_worlds}, our model shows comparable performance on R@1, and outperforms on MRR, R@5, R@10 and mean rank by 1.47\%, 3.47\%, 6\%, 5.08, respectively. Our model also outperforms CoAtt \cite{are_you_talking_to_me}, which achieved the previous best results for generative models. Our result surpass it with large margins on R@5, R@10 and mean rank by 3.06\%, 5.81\% and 5.28, respectively. 

While our model demonstrates remarkable improvement on R@5, R@10 and mean rank, MRR shows moderate gain while R@1 is slightly behind. We attribute this to the fact that there could be more than one correct response among the candidates while only one is provided as \textit{the} correct answer. As demonstrated by the examples of top-10 responses in Figure~\ref{fig:rank_example}, our model is capable of ranking multiple correct answers to higher places. However, the single human answer is not necessarily ranked the 1st, thus greatly affecting R@1. 

%%%%%%%%%%%%%%%%%%%%%%%%%%%%%%%%%%%
\begin{figure}
\centering
\includegraphics[trim=0 120 600 13, clip,width=0.8\columnwidth]{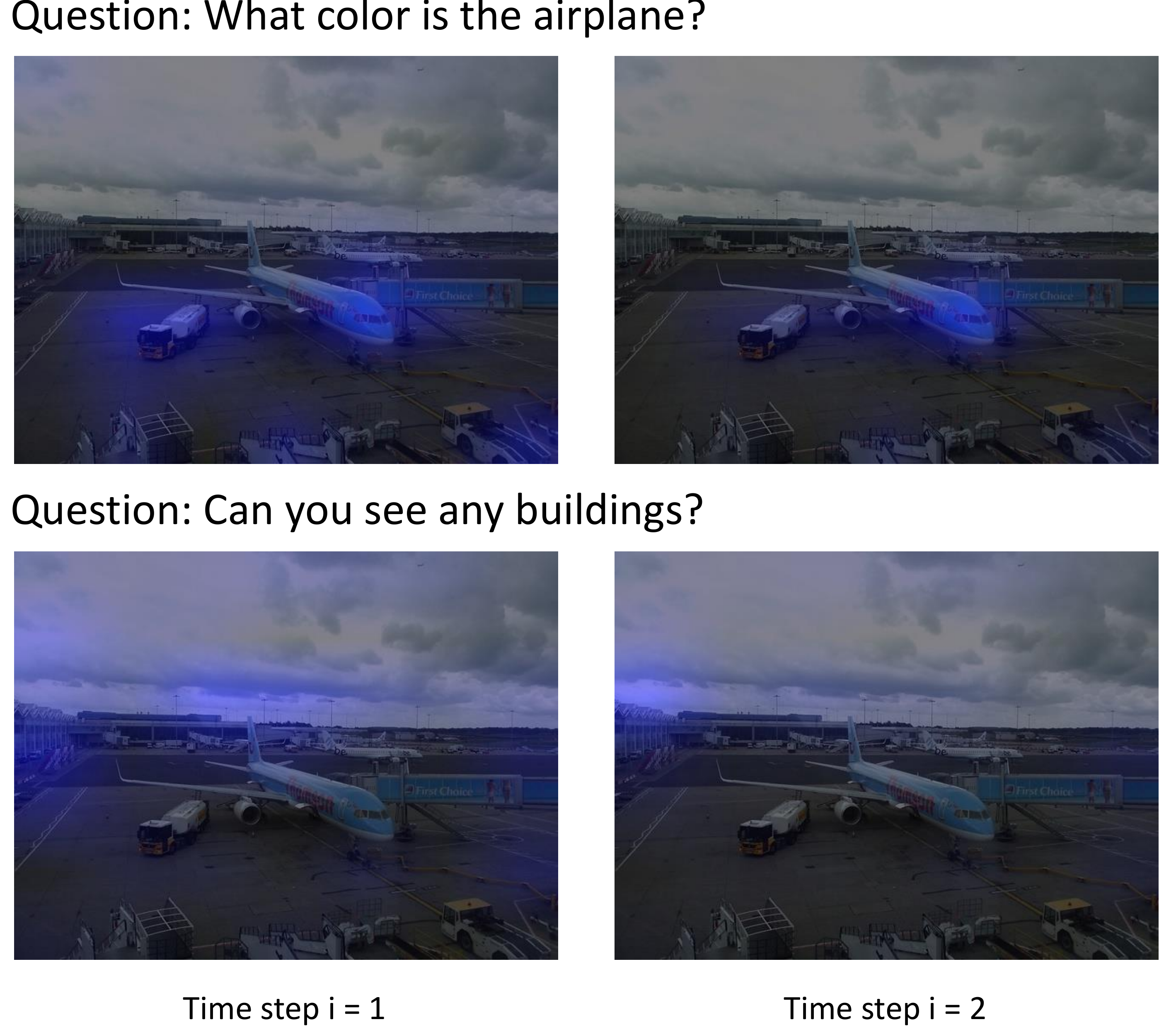}
\caption{Visualization of image attention heatmaps for different questions and reasoning steps. Regions of attention are highlighted in blue.}
\label{fig:att_viz}
\end{figure}
%%%%%%%%%%%%%%%%%%%%%%%%%%%%%%%%%%%

%%%%%%%%%%%%%%%%%%%%%%%%%%%%%
\begin{figure*}
\centering
\includegraphics[trim=70 274 340 60,clip,width=0.9\textwidth]{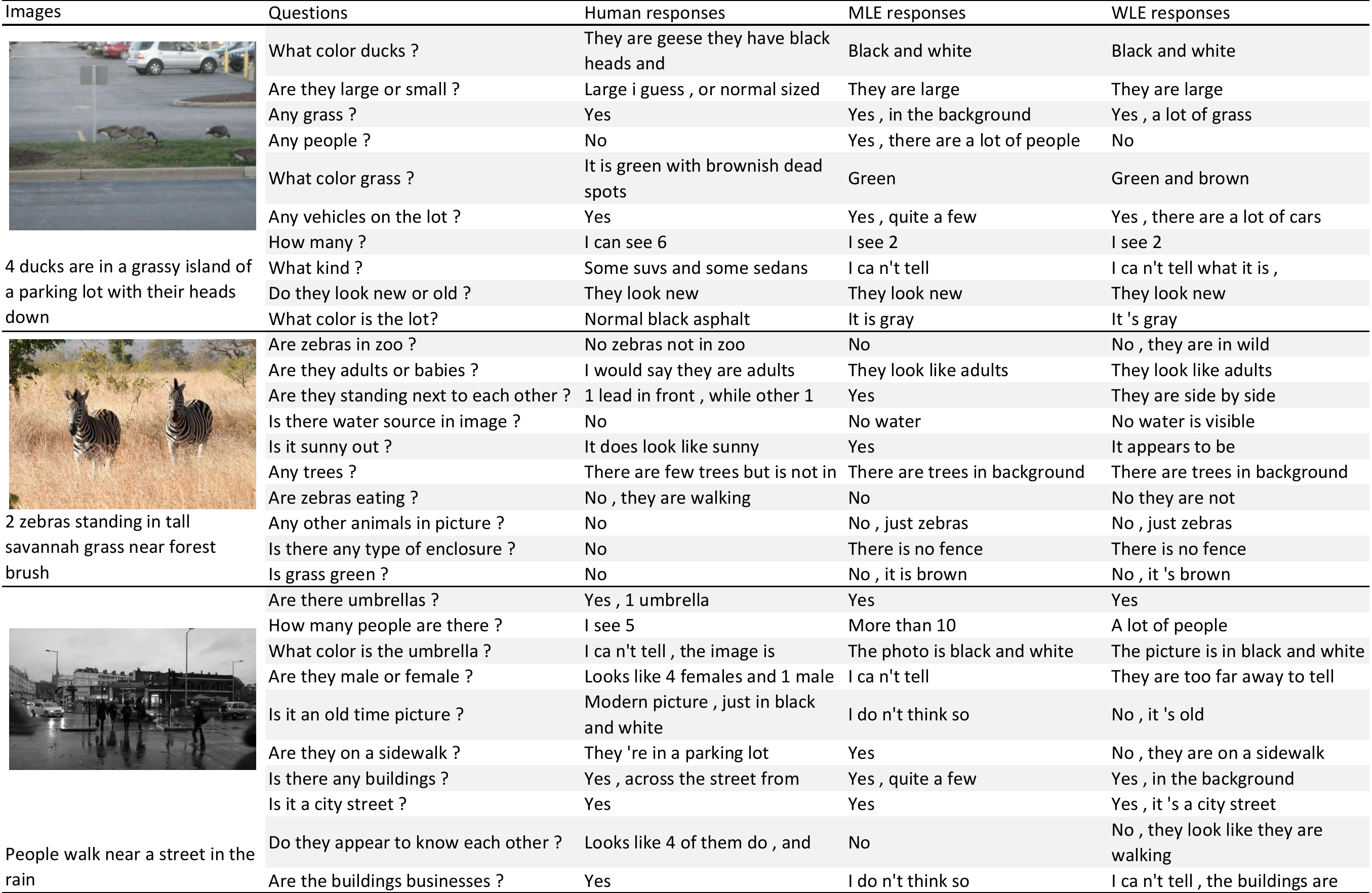}
\caption{Qualitative results on \texttt{test}. The questions and answers are truncated at 16 and 8, respectively, same as our data pre-processing. }
\label{fig:qualitative}
\end{figure*}
%%%%%%%%%%%%%%%%%%%%%%%%%%%%%

%\paragraph{Results on VisDial v1.0}
\subsubsection{Results on VisDial v1.0}
In the Visual Dialog Challenge 2018, all correct responses in \texttt{test} are annotated by humans and considered in the evaluation. Figure~\ref{fig:challenge} represents the top-10 results. Our model, as the only generative model in the top-10, ranked as the 6th among those discriminative models. It also verifies our claim that our low R@1 score on v0.9 is because the evaluation only considers the human response but ignore all other correct responses. We used ResNet features for the challenge.
Since ReDAN only reports its generative performance on VisDial v1.0 \texttt{val} with bottom-up features, we also present our results using the same setting in Table~\ref{tb:v1_0val}. We list the results of previous work in Table~\ref{tb:v1_0val} as reported in~\cite{ReDAN}. Similar to the results on VisDial v0.9, our proposed method outperforms previous methods on MRR, R@5, R@10 and Mean.

%%%%%%%%%%%%%%%%%%%%%%
%\input{fig_demo}
%%%%%%%%%%%%%%%%%%%%%%

%\paragraph{Ablation Study}
\subsubsection{Ablation Study}
Our model contains two main novel components, namely the adaptive multi-modal reasoning module and the WLE based training scheme. To verify the contribution of each component, we compare the following models %\footnote{We use the official source code of \cite{best_of_both_worlds} for HCIAE}
: (a) \textbf{HCIAE-MLE} is the HCIAE model trained via MLE; (b) \textbf{HCIAE-GAN} is the HCIAE model trained via MLE and GAN; (c) \textbf{HCIAE-WLE} is the HCIAE model trained via WLE; (d) \textbf{AMR-MLE} is our AMR model trained via MLE; (e) \textbf{AMR-WLE} is our final model with both key components.

The results are listed in Table \ref{tb:ablation_results}. The effectiveness of the proposed reasoning scheme is demonstrated in the HCIAE-MLE vs. AMR-MLE and HCIAE-WLE vs. AMR-WLE comparisons where our model outperforms HCIAE on all metrics. 
The importance of our proposed WLE is highlighted in the comparison between HCIAE-WLE and HCIAE-GAN. HCIAE-WLE performs better on all metrics. Specifically, the improvement on the HCIAE model by WLE is more than twice of that by GAN on R@10 (6.35 vs. 2.31) and mean rank (6.08 vs. 1.78). Our proposed training scheme is therefore also compatible and effective with other encoders.

%\paragraph{Qualitative Results}
\subsubsection{Qualitative Results}
Examples of image attention heatmaps are visualized in Figure~\ref{fig:att_viz}, which demonstrate the adaptive reasoning focuses for different questions and reasoning time steps. For example, for the second question, the attention on image was first at a large area of background, then moved to more focused region to answer the question 'any buildings'.

Figure~\ref{fig:qualitative} shows some qualitative results on \texttt{test}. Our generative model is able to generate more non-generic answers. As evidently shown in the comparison between MLE and WLE, the WLE results are more specific and human-like.

%We have built a demo of the Visual Dialog system, which takes input questions from a user and answers questions regarding an image. If the paper gets accepted for publication, we will be happy to release the demo (we cannot release it at this point due to author anonymity requirements). Figure~\ref{fig:demo} shows a screenshot from this demo.

\section{Conclusion}
In this work, we have presented a novel generative visual dialogue system. It involves an adaptive reasoning module for multi-modal inputs. The proposed reasoning module does not have any pre-defined sequential reasoning order and can accommodate various dialogue scenarios. The generative visual dialogue system is trained using weighted likelihood estimation, for which we design a new training scheme for generative visual dialogue systems.

\bibliographystyle{named}
\bibliography{ijcai19}

\end{document}